\documentclass[11pt,a4paper]{article}
\usepackage[hyperref]{acl2019}
\usepackage{times}
\usepackage{latexsym}

\usepackage{kotex}
\usepackage{multirow}
\usepackage{graphicx}
\usepackage[normalem]{ulem}
\usepackage{makecell}
\usepackage{marvosym}

\usepackage{url}

\aclfinalcopy 

\title{Extracting Arguments from Korean Question and Command:\\An Annotated Corpus for Structured Paraphrasing}

\author{Won Ik Cho$^\circ$, Young Ki Moon\textsuperscript{\Cross}, Woo Hyun Kang$^\circ$, Nam Soo Kim$^\circ$ \\
	Department of Electrical and Computer Engineering and INMC, Seoul National University$^\circ$\\
	{\tt \{wicho,whkang\}@hi.snu.ac.kr, nkim@snu.ac.kr}\\
	Department of Computer Engineering, Inha University\textsuperscript{\Cross}\\
	{\tt ykmoon0814@gmail.com} \\}

\date{}

\begin{document}
\maketitle
\begin{abstract}
  Intention identification is a core issue in dialog management. However, due to the non-canonicality of the spoken language, it is difficult to extract the content automatically from the conversation-style utterances. This is much more challenging for  languages like Korean and Japanese since the agglutination between morphemes make it difficult for the machines to parse the sentence and understand the intention. To suggest a guideline for this problem, and to merge the issue flexibly with the neural paraphrasing systems introduced recently, we propose a structured annotation scheme for Korean question/commands and the resulting corpus which are widely applicable to the field of argument mining. The scheme and dataset are expected to help machines understand the intention of natural language and grasp the core meaning of conversation-style instructions.
\end{abstract}

\section{Introduction}

In a semantic and pragmatic view, questions and commands differ from interrogatives and imperatives, respectively. We can easily observe the particular types of declaratives (1a,b) which explicitly require the addressee to give an answer or to take action. Also, some rhetorical questions (1c) and commands (1d) do not require a response.\smallskip\\
(1) a. \textit{I want to know why he keeps that hidden.}\\
\phantom{(1) }b. \textit{I think you should go now.}\\
\phantom{(1) }c. \textit{Why should you be surprised?}\\
\phantom{(1) }d. \textit{Imagine what it must've been like for them.}\smallskip

In identifying the intention and filling slots for conversational sentences, aforementioned characteristics make it difficult for the spoken language understanding systems to catch what the speaker intends. For these reasons, the concept of dialogue act \citep{stolcke2000dialogue,bunt2010towards} was introduced to categorize the sentences regarding the illocutionary act \citep{searle1976classification}, but such elaborate categorization may not fit with managing the core content of question/commands, or the \textit{arguments} as will be referred afterward. In this paper, discourse component \citep{portner2004semantics} is more investigated as a key in analyzing the instructional and directive utterances, for slot-filling and dialog management.

Here, we construct a criteria set on materializing arguments from non-rhetorical questions and commands, especially annotating corpus on Seoul Korean, which is a less explored language\footnote{For Korean text, word-level paraphrasing \citep{park2016affix} and news summarization \citep{jeong2016efficient} were suggested, but little was done on structured paraphrasing.}. 
The primary goal of the resource is to encourage a \textbf{flexible management of instructions} in a dialog with non-canonical input utterances. For instance, in the questions \textit{``Where should we go today''} and \textit{``Where is my wallet''}, the former asks for \textit{the destination} while the latter pursue \textit{the location} of the object. If the above kind of distinction is performed, the dialog system may better be able to deal with the users' needs in the circumstances regarding non-task-oriented conversation. The core component of the instructions can also be utilized in \textbf{making a plausible reaction}, such as \textit{``I will find you the destination''}.

\section{Corpus Annotation}

In this section, an annotation scheme regarding the patterns of questions and commands is described. Note that punctuations were omitted assuming ASR result as an input.
Briefly on the annotation scheme, for the sentences with (c)overt speech act (SA) layer of question/command, both extractive and abstractive paraphrasing are utilized depending on the content. Redundant functional particles were removed to guarantee that the output is an independent phrase. 

\subsection{Questions}

For each question, its argument and \textit{question type} were annotated. Here, questions include not only the interrogatives but also the declaratives with predicates such as \textit{want to know} or \textit{wonder}. In the annotation process, rhetorical questions \citep{rohde2006rhetorical} were excluded.

\textbf{Question type} was tagged with three labels, namely yes/no, alternative, and \textit{wh-}questions \citep{huddleston1994contrast}. \textbf{Yes/no} question, also known as polar question, has a possible answer set of yes/no (2a). \textbf{Alternative} question is the question which gives multiple candidates and requires a choice (2b). \textbf{\textit{Wh-}}question is the type of questions regarding \textit{wh-} particles, namely who, what, where, when, why, and how (2c-h).\smallskip\\
(2a) 너 \textbf{의료 봉사 신청} \uline{했어}\\
\phantom{(1a) }ne \textbf{uylyo pongsa sincheng} \uline{hayss-e}\\
\phantom{(1a) }you \textbf{medical service apply} \uline{did-INT}\\
\phantom{(1a) }\textit{Did you apply for medical service?}\smallskip\\
(2b) \textbf{버스}로 \textbf{올}\uline{거야} \textbf{택시}로 \textbf{올}\uline{거야}\\
\phantom{(1a) }\textbf{pesu}-lo \textbf{ol}-\uline{keya} \textbf{thayksi}-lo \textbf{ol}-\uline{keya}\\
\phantom{(1a) }\textbf{bus}-by \textbf{come}-\uline{INT} \textbf{taxi}-by \textbf{come}-\uline{INT}\\
\phantom{(1a) }\textit{Will you come by bus or taxi?}\smallskip\\
(2c) \textbf{오늘}은 \uline{누구} \textbf{왔}니\\
\phantom{(1a) }\textbf{onul}-un \uline{nwukwu} \textbf{wass}-ni\\
\phantom{(1a) }\textbf{today}-TOP \uline{who} \textbf{came}-INT\\
\phantom{(1a) }\textit{Who came today?}\smallskip\\
(2d) \textbf{스톡옵션}이 \uline{뭔} 줄 아니\\
\phantom{(1a) }\textbf{suthokopsyen}-i \uline{mwen} cwul a-ni\\
\phantom{(1a) }\textbf{stock-option}-NOM \uline{what} is.ACC know-INT\\
\phantom{(1a) }\textit{Do you know what stock option is?}\smallskip\\
(2e) \uline{어디} \textbf{있}니 로비야\\
\phantom{(1a) }\uline{eti} \textbf{iss}-ni Robi-ya\\
\phantom{(1a) }\uline{where} \textbf{be}-INT Robi-VOC\\
\phantom{(1a) }\textit{Where are you, Robi?}\smallskip\\
(2f) \textbf{대구} \uline{몇 시}에 \textbf{도착}이야\\
\phantom{(1a) }\textbf{taykwu} \uline{myech si}-ey \textbf{tochak}-iya\\
\phantom{(1a) }\textbf{Daegu} \uline{what hour}-TIM \textbf{arrival}-INT\\
\phantom{(1a) }\textit{When do you arrive in Daegu?}\smallskip\\
(2g) 이 동네 갑자기 \uline{왜} 이렇게 \textbf{막히}지 \\
\phantom{(1a) }i tongney kapcaki \uline{way} ileh-key \textbf{makhi}-ci\\
\phantom{(1a) }this town suddenly \uline{why} this-like \textbf{jam}-INT\\
\phantom{(1a) }\textit{Why is this town suddenly jammed like this?}\smallskip\\
(2h) \textbf{해외 송금} \uline{어떻게} 하는 거야\\
\phantom{(1a) }\textbf{hayoy songkum} \uline{ettehkey} hanun ke-ya\\
\phantom{(1a) }\textbf{aboard remittance} \uline{how} doing thing-INT\\
\phantom{(1a) }\textit{How can I send money abroad?}\smallskip\\
\textbf{Argument extraction from the questions} was done depending on the question type. For yes/no questions, the content was appended with the term `-(인)지 or 여부' ([-(in)ci] or [yepwu], both meaning \textit{whether or not}), to make up a nominalized term for the query (3a). For alternative questions (3b), all the items were sequentially arranged in the form of `(A B 중) -한/할 것' ([(\textit{A} \textit{B} cwung) -han/hal kes],
\textit{what is/to do - between A and B}). For various types of \textit{wh-}questions we tried to avoid repeating the \textit{wh-}particles in the extraction and instead used the \textit{wh-}related terms such as `사람' ([sa-lam], \textit{person}), `의미' ([uy-mi], \textit{meaning}), `위치' ([wi-chi], \textit{place}), `시간' ([si-kan], \textit{time}), `이유' ([i-yu], \textit{reason}), `방법' ([pang-pep], \textit{method}) to guarantee the structuredness of the extraction and the utility for further usages such as web searching (3c-h). The results below correspond with the sentences (2a-h).\smallskip\\
(3a) \textbf{의료 봉사 신청} \uline{여부}\\
\phantom{(1a) }\textbf{uylyo pongsa sincheng} \uline{yepwu}\\
\phantom{(1a) }\textbf{medical service apply} \uline{presence}\\
\phantom{(1a) }\textit{Whether or not applied to medical service}\smallskip\\
(3b) \textbf{버스 택시} \uline{중} \textbf{타고\footnote{타-/\textit{ride} is usually accompanied with the transportation.} 올} \uline{것}\\
\phantom{(1a) }\textbf{pesu thayksi} \uline{cwung} \textbf{tha-ko ol} \uline{kes}\\
\phantom{(1a) }\textbf{bus taxi} \uline{between} \textbf{ride-PRG come} \uline{thing}\\
\phantom{(1a) }\textit{What to ride between bus and taxi}\smallskip\\
(3c) \textbf{오늘 온} \uline{사람}\\
\phantom{(1a) }\textbf{onul on} \uline{salam}\\
\phantom{(1a) }\textbf{today came} \uline{person}\\
\phantom{(1a) }\textit{The person who came today}\smallskip\\
(3d) \textbf{스톡옵션} \uline{의미}\\
\phantom{(1a) }\textbf{suthokopsyen} \uline{uymi}\\
\phantom{(1a) }\textbf{stock-option} \uline{meaning}\\
\phantom{(1a) }\textit{The meaning of stock option}\smallskip\\
(3e) 지금 \textbf{있}는 \uline{위치}\\
\phantom{(1a) }cikum \textbf{iss}-nun \uline{wichi}\\
\phantom{(1a) }now \textbf{be}-PRG \uline{place}\\
\phantom{(1a) }\textit{The place currently belong to}\smallskip\\
(3f) \textbf{대구 도착} \uline{시간}\\
\phantom{(1a) }\textbf{taykwu tochak} \uline{sikan}\\
\phantom{(1a) }\textbf{Daegu arrival} \uline{time}\\
\phantom{(1a) }\textit{Arrival time for Daegu}\smallskip\\
(3g) \textbf{막히}는 \uline{이유}\\
\phantom{(1a) }\textbf{makhi}-nun \uline{iyu}\\
\phantom{(1a) }\textbf{jam}-PRG \uline{reason}\\
\phantom{(1a) }\textit{The reason for jam}\smallskip\\
(3h) \textbf{해외 송금} \uline{방법}\\
\phantom{(1a) }\textbf{hayoy songkum} \uline{pangpep}\\
\phantom{(1a) }\textbf{abroad remittance} \uline{method}\\
\phantom{(1a) }\textit{The way to send money abroad}\smallskip

\subsection{Commands}

For each command, its argument and the \textit{negativeness} were annotated. Here, commands include not only the imperative forms with covert subject and the requests in the interrogative form (different from the categorization in \citet{portner2004semantics}), but also the wishes and exhortatives that induce the addressee’s response. Imperatives used as exclamation or evocation are not included since they are considered rhetorical. The optatives that are used idiomatically, such as \textit{Have a nice day!} \citep{han2000structure}, are also not included since the feasibility of the to-do-lists is beyond the addressee’s capacity.

\textbf{Negativeness} was tagged with three labels, namely prohibitions, requirements, and strong requirements. \textbf{Prohibition (PH)} is the type of command that stops or prohibits an action. It possibly contains negations (4a1) or the predicates/modifiers that induce the prohibition (4a2). \textbf{Requirement (REQ)} is the type of command that is positive, with no terms that induce the restriction (4b1), and corresponds with various sentence forms aforementioned. The imperatives with information-seeking intent (4b2) are treated separately as a question. \textbf{Strong requirement (SR)} is the type of command where the prohibition and requirement are concatenated sequentially, appearing in spoken Korean as an emphasis (4c)\footnote{In English, the order is generally reversed, as in \textit{I told you to slay the dragon, not lay it}.}.\smallskip\\
(4a1) 태풍 오니까 \textbf{밖에 나가}지 \uline{마}\\
\phantom{(1a) }thayphwung o-nikka \textbf{pakk-ey naka}-ci \uline{ma}\\
\phantom{(1a) }typhoon come-because \textbf{outside-to go}-ci \uline{NEG}\\
\phantom{(1a) }\textit{Don't go outside, typhoon comes.}\smallskip\\
(4a2) \textbf{안전띠} \uline{안}\textbf{매}면 \uline{큰일나}\\
\phantom{(1a) }\textbf{ancentti} \uline{an}-\textbf{may}-myen \uline{khunil-na}\\
\phantom{(1a) }\textbf{seatbelt} \uline{no}-\textbf{take}-if \uline{danger-occur.DEC}\\
\phantom{(1a) }\textit{It's dangerous if you don't take a seatbelt.}\smallskip\\
(4b1)  \textbf{인적사항 확인} \uline{바랍니다}\\
\phantom{(1a) }\textbf{inceksahang hwakin} \uline{palap-nita}\\
\phantom{(1a) }\textbf{personal-info check} \uline{want-HON.DEC}\\
\phantom{(1a) }\textit{I want you to check the personal info.}\smallskip\\
(4b2) \textbf{이번 주 일정}을 \textbf{모두} \uline{말해}\\
\phantom{(1a) }\textbf{ipen cwu ilceng}-ul \textbf{motwu} \uline{mal-hay}\\
\phantom{(1a) }\textbf{this week schedule}-ACC \textbf{all} \uline{tell-IMP}\\
\phantom{(1a) }\textit{Tell me all the schedules this week.}\smallskip\\
(4c) 욕심부리지 \uline{말고} \textbf{지금 팔}\uline{아}\\
\phantom{(1a) }yoksim-pwuli-ci \uline{malko} \textbf{cikum phal}-\uline{a}\\
\phantom{(1a) }greedy-be-ci \uline{not-and} \textbf{now sell}-\uline{IMP} \\
\phantom{(1a) }\textit{Don't be greedy, just sell it now!}\smallskip\\
\begin{table}
	\centering
	\resizebox{0.9\columnwidth}{!}{%
		\begin{tabular}{|c|c|c|c|}
			\hline
			\multicolumn{1}{|l|}{}              & \multicolumn{2}{c|}{\textbf{Types}}                                                                & \textbf{Correspondings}                                                      \\ \hline
			\multirow{8}{*}{\textbf{Questions}} & \multicolumn{2}{c|}{\textit{Yes/no}}                                                               & \begin{tabular}[c]{@{}c@{}}whether or not\\ -(인)지, 여부\end{tabular}        \\ \cline{2-4} 
			& \multicolumn{2}{c|}{\textit{Alternative}}                                                          & \begin{tabular}[c]{@{}c@{}}what is/to do between\\ -랑 -중 -한/할 것\end{tabular} \\ \cline{2-4} 
			& \multirow{6}{*}{\textit{\begin{tabular}[c]{@{}c@{}}Wh-\\ questions\end{tabular}}} & \textit{Who}   & \begin{tabular}[c]{@{}c@{}}person, identity\\ 사람, 정체\end{tabular}            \\ \cline{3-4} 
			&                                                                                   & \textit{What}  & \begin{tabular}[c]{@{}c@{}}meaning\\ 의미\end{tabular}                         \\ \cline{3-4} 
			&                                                                                   & \textit{Where} & \begin{tabular}[c]{@{}c@{}}location, place \\ 위치, 장소\end{tabular}            \\ \cline{3-4} 
			&                                                                                   & \textit{When}  & \begin{tabular}[c]{@{}c@{}}time, period, hour\\ 시간, 기간, 시각\end{tabular}      \\ \cline{3-4} 
			&                                                                                   & \textit{Why}   & \begin{tabular}[c]{@{}c@{}}reason\\ 이유\end{tabular}                          \\ \cline{3-4} 
			&                                                                                   & \textit{How}   & \begin{tabular}[c]{@{}c@{}}method, measure\\ 방법, 대책\end{tabular}             \\ \hline
			\multirow{3}{*}{\textbf{Commands}}  & \multicolumn{2}{c|}{\textit{Prohibitions}}                                                         & \begin{tabular}[c]{@{}c@{}}Prohibition: not to -\\ -지 않기 (금지)\end{tabular}          \\ \cline{2-4} 
			& \multicolumn{2}{c|}{\textit{Requirements}}                                                         & \begin{tabular}[c]{@{}c@{}}Requirement: -ing\\ -기 (요구)\end{tabular}          \\ \cline{2-4} 
			& \multicolumn{2}{c|}{\textit{\begin{tabular}[c]{@{}c@{}}Strong\\ Requirements\end{tabular}}}        & \begin{tabular}[c]{@{}c@{}}Requirement: -ing\\ -기 (요구)\end{tabular}          \\ \hline
		\end{tabular}%
	}
	\caption{Structured annotation scheme.}
	\label{my-label}
\end{table}
\textbf{Argument extraction from the commands} was done with a nominalized predicate ‘-하(지 않)기' ([-ha-(ci ahn)-ki], \textit{doing (not to do) something}). For PH, the action that is prohibited is annotated with a negation (5a1). For REQ, the requirement is annotated (5b1), and information-seeking ones are dealt as questions (5b2). For SR we only annotated the action that is required (5c), for disambiguation and an effective representation of a to-do-list. \smallskip\\
(5a1) \textbf{밖에 나가}지 않기 (금지)\\
\phantom{(1a) }\textbf{pakk-ey naka}-ci anh-ki\\
\phantom{(1a) }\textbf{outside-to go}-ci not-NMN\footnote{Denotes a nominalizer.}\\
\phantom{(1a) }\textit{Not to go outside} (prohibition)\smallskip\\
(5a2) \textbf{안전띠 매}기 (요구) \\
\phantom{(1a) }\textbf{ancentti may}-ki\\
\phantom{(1a) }\textbf{seatbelt take}-NMN\\
\phantom{(1a) }\textit{Taking a seatbelt} (requirement)\smallskip\\
(5b1) \textbf{인적사항 확인}하기 (요구)\\
\phantom{(1a) }\textbf{inceksahang hwakin}-haki\\
\phantom{(1a) }\textbf{personal info check}-NMN\\
\phantom{(1a) }\textit{Checking the personal info} (requirement)\smallskip\\
(5b2) \textbf{이번 주 모든 일정}\\
\phantom{(1a) }\textbf{ipen cwu motun ilceng}\\
\phantom{(1a) }\textbf{this week all schedule}\\
\phantom{(1a) }\textit{The schedule of this week} (\textit{wh-} question)\smallskip\\
(5c) \textbf{지금 팔}기 (요구) \\
\phantom{(1a) }\textbf{cikum phal}-ki\\
\phantom{(1a) }\textbf{now sell}-NMN \\
\phantom{(1a) }\textit{Selling it now} (requirement)\smallskip

There are points to be clarified regarding (4a2) and (5a2). Although (4a2) displays a property of prohibition induced by `큰일나 (danger occurs)', the target action contains a negation `안' which induces a double negation. Therefore, (5a2) was labeled as SR.

Since the commands hardly accompany abstract concept as \textit{wh-}questions do, the arguments were obtained mostly in an extractive way. Also, since the command inevitably includes a detailed to-do-list, the removal of functional particles was done only if they were considered redundant, unlike it was highly recommended for the questions.

\begin{table}
	\centering
	\resizebox{0.9\columnwidth}{!}{%
		\begin{tabular}{|c|c|c|}
			\hline
			& \textbf{Types} & \textbf{Portion} \\ \hline
			\multirow{3}{*}{\textbf{\begin{tabular}[c]{@{}c@{}}Questions\\ (17,869)\end{tabular}}} & \textit{Yes/no} & 5,718 (31.99\%) \\ \cline{2-3} 
			& \textit{Alternative} & 227 (1.27\%) \\ \cline{2-3} 
			& \textit{Wh- question} & 11,924 (66.73\%) \\ \hline
			\multirow{3}{*}{\textbf{\begin{tabular}[c]{@{}c@{}}Commands\\ (12,968)\end{tabular}}} & \textit{Prohibition} & 477 (3.67\%) \\ \cline{2-3} 
			& \textit{Requirement} & 12,369 (95.38\%) \\ \cline{2-3} 
			& \textit{\begin{tabular}[c]{@{}c@{}}Strong\\ requirement\end{tabular}} & 122 (0.94\%) \\ \hline
		\end{tabular}%
	}
	\caption{Dataset specification, denoted with the number of instances for each category and the portion. In the disclosed dataset, six types of sentences are randomly distributed, with the labels 0 to 5 in the order stated in the table.}
	\label{my-label}
\end{table}

\section{Dataset Specification}

We adopted the spoken Korean dataset of size 800K which was primarily constructed for language modeling and speech recognition of Korean. The sentences are in conversation-style and partly non-canonical, and the content covers topics such as weather, news, housework, e-mail, and stock. From the corpus we randomly selected 20K sentences and classified them into seven sentence types: fragments, 
rhetorical questions, rhetorical commands, questions, commands, and statements, with the inter-annotator agreement (IAA) of $\kappa$ = 0.85 \citep{fleiss1971measuring}. Questions and commands were chosen among them, and later, additional sets of questions and commands were augmented via manual generation, to make up the whole dataset of size 30,837.

The specification of the annotated corpus is displayed in Table 2. Since the annotation is quite explicitly defined for both question and command in view of discourse component \citep{portner2004semantics}, we performed a double-check instead of finding out a separate IAA. 

Due to the characteristics of the adopted corpus as a spoken language script targeting smart home agents, the portion of straight questions and commands (yes/no$\cdot$\textit{wh-}questions and REQ) is much higher than in the real-life language. We observed that the alternative questions, PH, and SR (especially the scrambled order and double negation) are relatively scarce compared with the portion within the human conversation, which will be augmented via crowd-sourcing in the future work. 

\section{Conclusion}
\label{sec:length}

In this paper, we proposed a structured annotation scheme for the argument extraction of conversation-style Korean questions and commands, concerning the discourse component they show. This is the first dataset on question set/to-do-list extraction for spoken Korean, up to our knowledge, and we annotated the syntax-related properties for the potential usage. This study may provide an appropriate guideline that helps extract an argument from the various non-canonical type of instructions in real life.  

Despite the small volume, the dataset incorporates consistency in the way it was constructed. Thus, in case of need, utterance-argument pairs can be created without difficulty referring to the examples and be augmented to the original corpus. Moreover, for easier construction, merely some arguments can be provided to the participants so that they create conversation-style question/commands with no regulation (e.g., generating ``\textit{Tell me where my phone is}'' from `\textit{the location of the speaker's phone}'). Here, the arguments that are frequently exploited in AI services can be adopted so as to boost the industrial utility of the corpus. 

In the aspect of linguistic characteristics, the annotation scheme can be extended to the languages that are morphologically rich and syntactically similar to Korean, such as Japanese. Expansion to other languages such as English, by utilizing the terms \textit{if-}, \textit{whether-}, or \textit{the place/reason, etc.}, is also expected to be available, though the methodology may be less impactful than in the aforementioned languages. Nevertheless, the scheme can be adopted by the languages where the act of question/command presents, and fits well with the spoken language analysis flourishing with the smart agents widely used nowadays. The dataset and scheme are freely available on-line\footnote{https://github.com/warnikchow/sae4k}.

%

\bibliography{my_bib_190624}

\begin{thebibliography}{10}
\expandafter\ifx\csname natexlab\endcsname\relax\def\natexlab#1{#1}\fi

\bibitem[{Bunt et~al.(2010)Bunt, Alexandersson, Carletta, Choe, Fang, Hasida,
  Lee, Petukhova, Popescu-Belis, Romary et~al.}]{bunt2010towards}
Harry Bunt, Jan Alexandersson, Jean Carletta, Jae-Woong Choe, Alex~Chengyu
  Fang, Koiti Hasida, Kiyong Lee, Volha Petukhova, Andrei Popescu-Belis,
  Laurent Romary, et~al. 2010.
\newblock Towards an iso standard for dialogue act annotation.
\newblock In \emph{Seventh conference on International Language Resources and
  Evaluation (LREC'10)}.

\bibitem[{Fleiss(1971)}]{fleiss1971measuring}
Joseph~L Fleiss. 1971.
\newblock Measuring nominal scale agreement among many raters.
\newblock \emph{Psychological bulletin}, 76(5):378.

\bibitem[{Han(2000)}]{han2000structure}
Chung-hye Han. 2000.
\newblock \emph{The structure and interpretation of imperatives: mood and force
  in Universal Grammar}.
\newblock Psychology Press.

\bibitem[{Huddleston(1994)}]{huddleston1994contrast}
Rodney Huddleston. 1994.
\newblock The contrast between interrogatives and questions.
\newblock \emph{Journal of Linguistics}, 30(2):411--439.

\bibitem[{Jeong et~al.(2016)Jeong, Ko, and Seo}]{jeong2016efficient}
Hyoungil Jeong, Youngjoong Ko, and Jungyun Seo. 2016.
\newblock Efficient keyword extraction and text summarization for reading
  articles on smart phone.
\newblock \emph{Computing and Informatics}, 34(4):779--794.

\bibitem[{Park et~al.(2016)Park, Gweon, and Heo}]{park2016affix}
Hancheol Park, Gahgene Gweon, and Jeong Heo. 2016.
\newblock Affix modification-based bilingual pivoting method for paraphrase
  extraction in agglutinative languages.
\newblock In \emph{Big Data and Smart Computing (BigComp), 2016 International
  Conference on}, pages 199--206. IEEE.

\bibitem[{Portner(2004)}]{portner2004semantics}
Paul Portner. 2004.
\newblock The semantics of imperatives within a theory of clause types.
\newblock In \emph{Semantics and linguistic theory}, volume~14, pages 235--252.

\bibitem[{Rohde(2006)}]{rohde2006rhetorical}
Hannah Rohde. 2006.
\newblock Rhetorical questions as redundant interrogatives.

\bibitem[{Searle(1976)}]{searle1976classification}
John~R Searle. 1976.
\newblock A classification of illocutionary acts.
\newblock \emph{Language in society}, 5(1):1--23.

\bibitem[{Stolcke et~al.(2000)Stolcke, Ries, Coccaro, Shriberg, Bates,
  Jurafsky, Taylor, Martin, Ess-Dykema, and Meteer}]{stolcke2000dialogue}
Andreas Stolcke, Klaus Ries, Noah Coccaro, Elizabeth Shriberg, Rebecca Bates,
  Daniel Jurafsky, Paul Taylor, Rachel Martin, Carol~Van Ess-Dykema, and Marie
  Meteer. 2000.
\newblock Dialogue act modeling for automatic tagging and recognition of
  conversational speech.
\newblock \emph{Computational linguistics}, 26(3):339--373.

\end{thebibliography}
\bibliographystyle{acl_natbib}

\end{document}